\documentclass[10pt,twocolumn,letterpaper]{article}

\usepackage{cvpr} 
\usepackage{times}
\usepackage{epsfig}
\usepackage{graphicx}
\usepackage{amsmath}
\usepackage{amssymb}

\usepackage{enumitem}
\usepackage{graphics} 
\usepackage{etoolbox}
\usepackage[font=small,labelfont=bf]{caption}
\usepackage{algorithm}
\usepackage{algorithmic}

\usepackage{booktabs}
\usepackage{multirow}
\usepackage{multicol}

\begin{document}

\title{MTCAE-DFER: Multi-Task Cascaded Autoencoder for Dynamic Facial Expression Recognition}

\author{Peihao Xiang, Kaida Wu, Ou Bai\\
Florida International University\\
Miami, Florida, USA\\
{\tt\small \{pxian001, kwu020, obai\}@fiu.edu}
}

\maketitle
\thispagestyle{empty}

\begin{abstract}

This paper expands the cascaded network branch of the autoencoder-based multi-task learning (MTL) framework for dynamic facial expression recognition, namely Multi-Task Cascaded Autoencoder for Dynamic Facial Expression Recognition (MTCAE-DFER). MTCAE-DFER builds a plug-and-play cascaded decoder module, which is based on the Vision Transformer (ViT) architecture and employs the decoder concept of Transformer to reconstruct the multi-head attention module. The decoder output from the previous task serves as the query (Q), representing local dynamic features, while the Video Masked Autoencoder (VideoMAE) shared encoder output acts as both the key (K) and value (V), representing global dynamic features. This setup facilitates the interaction between global and local dynamic features across related tasks. Additionally, this proposal aims to alleviate overfitting of complex large model. We utilize autoencoder-based multi-task cascaded learning approach to explore the impact of dynamic face detection and dynamic face landmark on dynamic facial expression recognition, which enhances the model’s generalization ability. After we conduct extensive ablation experiments and comparison with state-of-the-art (SOTA) methods on various public datasets for dynamic facial expression recognition, the robustness of the MTCAE-DFER model and the effectiveness of global-local dynamic feature interaction among related tasks have been proven.

\end{abstract}

\section{INTRODUCTION}

\begin{figure*}[t]
	\centering
	\includegraphics[scale=0.313]{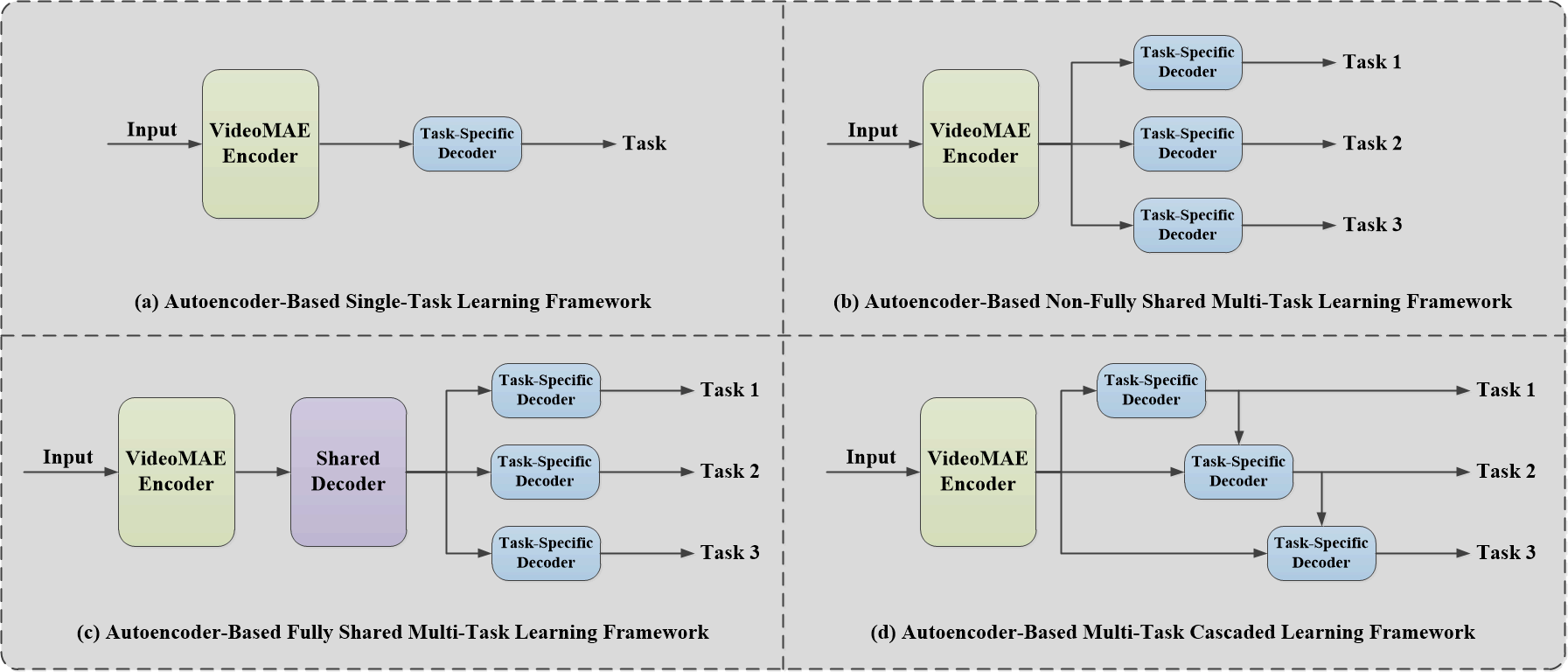}
	\caption{Illustration of the differences between the following four frameworks: (a) Autoencoder-Based Single-Task Learning Framework, (b) Autoencoder-Based Non-Fully Shared Multi-Task Learning Framework, (c) Autoencoder-Based Fully Shared Multi-Task Learning Framework and (d) Our Autoencoder-Based Multi-Task Cascaded Learning Framework.}
	\label{fig:framework}
\end{figure*}

Dynamic Facial Expression Recognition (DFER) is a crucial technology in human-computer interaction and affective computing, building on the foundations of Static Facial Expression Recognition (SFER) and further advancing its development. In SFER technology, most recognition models are static, serial, and context-free, such as DeepEmotion \cite{c1}, FER-VT \cite{c2}, and PAtt-Lite \cite{c3}. The preliminary stages of DFER development, the feasibility of three-dimensional convolutional neural network (3DCNN) was considered, with examples like C3D \cite{c4}, I3D \cite{c5}, and SlowFast \cite{c6}. However, these networks require significant computational power and still unable to meet real-time. With the emergence of sequence models such as RNN \cite{c7}, GRU \cite{c8}, and LSTM \cite{c9}, combining convolutional neural network with recurrent neural network has become a common trend. For example, models like ResNet18 + LSTM \cite{c10} and C3D + LSTM \cite{c11} have been developed to construct DFER system. Moreover, since the introduction of the TimeSformer \cite{c12} architecture, it is known for strengthening spatiotemporal contextual relevance, while more suitable for dynamic data. Examples include Former-DFER \cite{c13}, STT-DFER \cite{c14}, LOGO-Former \cite{c15}, and MAE-DFER \cite{c16}. However, since this architecture emphasizes the effectiveness of global feature extraction, it tends to overlook the key task-specific features. In other words, most current DFER models focus on global feature extraction and inference but lack attention to local feature and global-local interaction feature. This can lead to the model paying too much attention to broad features during the learning process, while ignoring the local features for the specific task.

To address the challenges mentioned above, although models like LOGO-Former \cite{c15} and MAE-DFER \cite{c16} have made some improvements, they still lack advanced technologies to further alleviate these shortcomings. Inspired by MTFormer \cite{c17} and MNC \cite{c18}, we propose using the multi-task cascaded learning approach to enhance the model robustness and the effectiveness of global-local feature interaction in DFER. As shown in Fig.1, (a) is Autoencoder-Based Single-Task Learning (STL) network, where the Video Masked Autoencoder (VideoMAE) \cite{c19} encoder is obtained by self-supervised learning. It uses the Masked Autoencoder (MAE) \cite{c20} method, but this network structure has weak generalization ability. (b) is Autoencoder-Based Non-Fully Shared Multi-Task Learning (MTL) network. The VideoMAE \cite{c19} encoder is shared, while non-shared task-specific decoders are designed for different tasks, improving the model's overall generalization ability through multiple tasks. (c) is Autoencoder-Based Fully Shared MTL network. It has a shared encoder-decoder pair and the task-specific decoder for each task, enhancing global feature interaction. (d) is Autoencoder-Based Multi-Task Cascaded Learning network. It employs the shared VideoMAE \cite{c19} encoder to obtain global representation information, and then uses the cascaded network to enable feature interaction across tasks, achieving global-local feature interaction of related tasks.

Combined with the aforementioned solutions, our purpose is to explore the impact of dynamic face detection and dynamic face landmark on DFER within the MTL framework, as well as the effectiveness of global-local feature interaction between related face tasks in the cascaded network. Therefore, we will propose a new framework of Multi-Task Cascaded Autoencoders for Dynamic Facial Expression Recognition, named MTCAE-DFER. The contributions of this work are as follows:

\begin{itemize}[itemsep=0em,topsep=0em]
  \item We build a multi-task cascaded autoencoder framework for DFER, extending the MTL branch based on the Transformer \cite{c21}. Furthermore, this framework can better play the parallel, dynamic and contextual of the DFER model, especially in human-computer interaction system, reducing the space-time complexity.
  \item We develop a plug-and-play cascaded decoder. To enable feature interaction between related tasks, the decoder is designed using the Transformer \cite{c21} decoder concept and the Vision Transformer (ViT) \cite{c22} architecture to facilitate global-local feature interaction between cascaded tasks.
  \item We conduct extensive experiments on three public dynamic facial expression datasets (RAVDESS \cite{c23}, CREMA-D \cite{c24}, MEAD \cite{c25}) to verify the competitiveness of MTCAE-DFER. Through ablation studies and comparisons with state-of-the-art (SOTA) methods, the generalization ability of the MTCAE-DFER model in DFER task and the effectiveness of global-local feature interaction between related tasks will be demonstrated.
\end{itemize}

\section{RELATED WORK}

\begin{figure*}[t]
	\centering
	\includegraphics[scale=0.23]{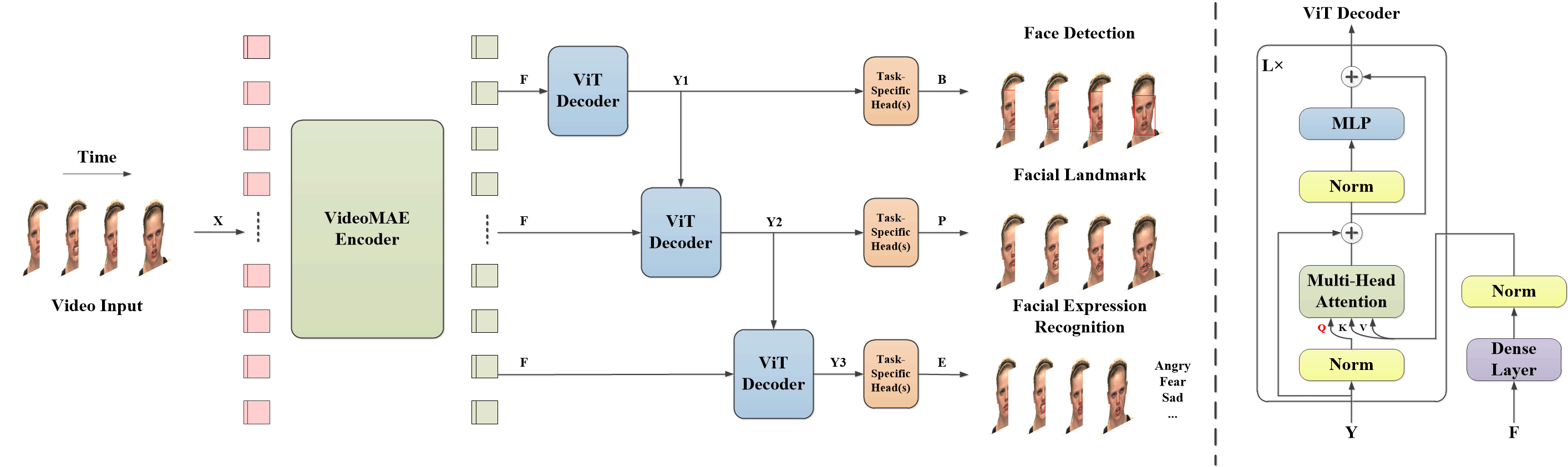}
	\caption{MTCAE-DFER Model Structure}
	\label{fig:structure}
\end{figure*}

\subsection{Dynamic Facial Expression Recognition}

DFER is the key part of dynamic emotion recognition, which focuses on the analysis and computation of facial emotions in visual data. At present, most DFER models use spatiotemporal attention as their core and built on the ViT \cite{c22} architecture, to capture contextual global features related to dynamic facial expressions and improve recognition accuracy. Examples include models like LOGO-Former \cite{c15} and MAE-DFER \cite{c16}. The LOGO-Former \cite{c15} architecture introduces the global-local feature interaction mechanism, which interacts global and local features from the spatio-temporal level, enhancing the separability of different data attributes in feature space. Additionally, the MAE-DFER \cite{c16} architecture is one of SOTA models in DFER. It employs the VideoMAE \cite{c19} self-supervised learning method to address data scarcity in deep learning, thereby producing the pre-trained model with strong feature extraction capabilities. In this work, we will adopt global-local feature interaction concept and inherit this VideoMAE \cite{c19} encoder pre-trained model. This approach will extract global feature of dynamic facial expressions, while strengthening the separability of local key feature from different related tasks level.

\subsection{Multi-Task Learning}

MTL is mainly used for joint learning of multiple tasks to enhance the generalization ability of the model. Various MTL variants have been developed by designing the shared module to effectively distribute and share information across different tasks, as seen in structures like MTFormer \cite{c17}. In this study, we focus on exploring and extending the various MTL frameworks shown in Fig. 1. Since this paper primarily deals with DFER, the tasks chosen for MTL will all be facial-related. Our research reveals that in traditional approaches to DFER from video, multiple models are often used in sequence: first to detect the face in the video, then to detect facial landmarks, and finally to recognize facial expressions. However, we aim to streamline the process by constructing the end-to-end unified model that simultaneously performs these three tasks: dynamic face detection, dynamic face landmark, and DFER.

\subsection{Cascaded Autoencoder}

CAE is the extension of the traditional autoencoder on cascaded network. Generally, the autoencoder consists of an encoder-decoder pair, where the encoder acts as the feature extractor in neural network architecture, and the decoder serves as the feature reasoner. Referring to the cascade concept of \cite{c18}, the relationships between different related tasks can be captured through cascaded decoder. This architecture enhances the performance of MTL network by facilitating global-local feature interaction across related tasks, which helps focus on the critical features necessary for solving the tasks. In this paper, we adopt autoencoder-based cascaded multi-task network architecture. The cascaded decoder enable interaction between the global features extracted by the shared encoder and the local features derived from the task-specific decoder. This framework will be used to prove the effectiveness of global-local feature interaction across different related tasks, thereby improving the robustness of the overall model.

\section{METHOD}

\subsection{Revisiting VideoMAE}

VideoMAE \cite{c19} is a significant advancement in self-supervised learning within the field of dynamic vision. It learns dynamic visual representation information by applying some pixel patches with tube masking over consecutive video frames and then using the autoencoder to reconstruct these masked patches. The VideoMAE \cite{c19} structure consists of a pair of asymmetric encoder-decoder, and the core theory is derived from ImageMAE \cite{c20}. Their basic backbone network is ViT \cite{c22}, which leverages the spatiotemporal attention module to capture the long short-term dependencies between dynamic features. According to our analysis, the VideoMAE \cite{c19} encoder extracts dynamic features from video information, while the decoder performs inference on these dynamic features to reconstruct those masked pixel patches. 

In this work, we directly inherit the VideoMAE \cite{c19} encoder as the shared feature extractor. As shown in Figure 2, the VideoMAE \cite{c19} encoder is used as the pre-trained model to obtain dynamic representation information. The input size is $ \textbf{X} \in {R}^{16 \times 224 \times 224 \times 3} $, representing the frame count, image height, image width, and image channels, respectively. In addition, the kernel size of the 3D patch is $2 \times 16 \times 16 \times 3 $. After patching, the sequence matrix $ \textbf{S} \in {R}^{1568 \times 1536}$ can be obtained. Since the embedding layer dimension is 1024, which matrix is reduced to $\hat{\textbf{X}} \in {R}^{1568 \times 1024}$ token sequences. Through the operation of the pre-trained encoder model, we can obtain the global dynamic representation information about the video, and its output size is $\textbf{F} \in {R}^{1568 \times 1024}$. The mathematical expressions for the above process are as follows:
\begin{equation}
\textbf{S} = \text{Patching}(\textbf{X})
\label{eq:patching}
\end{equation}
\begin{equation}
\hat{\textbf{X}} = \text{Embedding}(\textbf{S})
\label{eq:embedding}
\end{equation}
\begin{equation}
\textbf{F} = \text{Encoder}(\hat{\textbf{X}})
\label{eq:encoder}
\end{equation}

\begin{algorithm}[tb]
\caption{ViT Decoder Algorithm}
\label{alg:algorithm}
\textbf{Input}: $Y, F$ \\
\textbf{Parameter}: $L$ \\
\textbf{Output}: $Y$
\begin{algorithmic}[1] 
\STATE Let $l = 1$.
\STATE $K, V = \text{Norm}(\text{Dense}(F)) $
\WHILE{$l \leqslant L $}
\STATE $Q = \text{Norm}(Y) $
\STATE $Z = \text{MHA}(Q,K,V) $
\STATE $Z' = Z + Y $
\STATE $Y' = \text{MLP}(\text{Norm}(Z')) $
\STATE $Y = Y' + Z' $
\ENDWHILE
\STATE \textbf{return} $Y$
\end{algorithmic}
\end{algorithm}

\subsection{Cascaded ViT Decoder Module Design}

The Cascaded ViT Decoder is a plug-and-play module based on ViT \cite{c22} as the basic architecture and uses the Transformer \cite{c21} decoder concept to reconstruct the multi-head attention (MHA), as illustrated in Fig. 2 ViT Decoder. The focus is on the attention mechanism, where the query ($Q$), key ($K$), and value ($V$) inputs aren't sourced from the same input. Specifically, $Q$ is the normalized output from the previous stage, while $K$ and $V$ are the normalized dense outputs from the VideoMAE \cite{c19} encoder. Furthermore, as shown in Algorithm 1, the ViT Decoder has recursive algorithm of the built-in module, which the values of $K$ and $V$ remain constant, while $Q$ continuously updates throughout the recursion. $Q$, $K$, and $V$ are then processed through the MHA mechanism to generate the interaction features $Z$. Once the interaction features $Z$ are computed, the next operations follow the standard ViT architecture. Specifically, $Z$ undergoes residual connection $Y$ to produce $Z'$, followed by normalization and multi-layer perceptron (MLP) operations to obtain $Y'$. Finally, another residual connection between $Z'$ and $Y'$ completes the recursive output $Y$.

This module leverages MHA to directly associate feature maps from different feature layers, generating crucial feature attention map. More importantly, for facilitating the interaction between global and local dynamic features, the design of this module ensures that the feature matrix output from the VideoMAE \cite{c19} encoder serves as the global dynamic representation information. To prevent the forgetting of original global feature relationships, the form of $K$ and $V$ matrix is input into the attention layer, while the local representation information is input as the form of $Q$ matrix. In the cascaded mode, the $Q$ matrix is sourced from the output of the previous stage. In the recursive mode, the $Q$ matrix comes from the recursive computation output. In this setup, the shared feature matrix $K$ and $V$ represent global dynamic features and remain constant throughout the process. By associating them with the local dynamic features matrix $Q$, the process promotes interaction between global and local dynamic features.

In this work, the MTL framework employs the cascaded learning approach to accomplish three tasks: dynamic face detection, dynamic face landmark, and DFER tasks. As illustrated in Fig. 2, each task is equipped with the Cascaded ViT Decoder module to explore the effectiveness of global-local feature interaction across different related tasks. Based on this framework, we propose the following hypotheses for each task stage:

\begin{itemize}[itemsep=0em,topsep=0em]
  \item \textbf{First Task Stage (Dynamic Face Detection):} The hypothesis is to identify the primary facial target features and eliminate non-target spatial noise from the feature dimensions. This stage focuses on filtering out irrelevant features to ensure the robustness of the following tasks.
  \item \textbf{Second Task Stage (Dynamic Face Landmark):} The hypothesis is to locate critical facial landmarks, reducing interference from other facial features that are less relevant to the subsequent expression recognition task.
  \item \textbf{Third Task Stage (DFER):} The hypothesis involves segmenting the expression feature information space based on both global dynamic features and key local facial features. The goal is to enhance expression recognition by combining fine-grained local information (such as the movements of specific facial landmarks) with overarching global context.
\end{itemize}

In short, by incorporating the global-local feature interaction at each task level to strengthen the attention of fine-grained local features, so that the information representation space is generalized.

\subsection{MTCAE-DFER: Model Structure}

As shown in Fig. 2, the MTCAE-DFER model mainly consists of VideoMAE \cite{c19} encoder, Cascaded ViT Decoder(s) and Task-Specific Head(s). This model directly takes video as input, and its input size is $ \textbf{X} \in {R}^{16 \times 224 \times 224 \times 3} $. As mentioned in Section 3.1, the VideoMAE \cite{c19} encoder serves as the shared feature extractor for all tasks, which generates global dynamic features $\textbf{F} \in {R}^{1568 \times 1024}$. In addition, three face-related tasks are addressed using the multi-task cascaded learning approach, where three ViT Decoder modules with feature dimension of 512 are constructed to produce three different local dynamic feature outputs $\textbf{Y} \in {R}^{1568 \times 512}$. Each ViT Decoder for these tasks uses $\textbf{F}$ as the shared global feature input, which is fed into $K$ and $V$ to stand for global representation information. On the other hand, the local feature input $Q$ of each task is different. Except for the first dynamic face detection task, which uses $\textbf{F}$ as local representation information. The other tasks use the ViT Decoder output $\textbf{Y}$ of the previous task as local representation information. For example, $\textbf{Y1}$ serves as the local feature input for the second task, and $\textbf{Y2}$ serves as the local feature input for the third task as shown in Fig. 2. The mathematical expressions for the above process are as follows:
\begin{equation}
\textbf{Y1} = \text{ViT Decoder}(\textbf{F},\textbf{F})
\label{eq:decoder1}
\end{equation}
\begin{equation}
\textbf{Y2} = \text{ViT Decoder}(\textbf{Y1},\textbf{F})
\label{eq:decoder2}
\end{equation}
\begin{equation}
\textbf{Y3} = \text{ViT Decoder}(\textbf{Y2},\textbf{F})
\label{eq:decoder3}
\end{equation}

Finally, the ViT Decoder output $\textbf{Y}$ of each task is input to the corresponding Task-Specific Head. The Task-Specific Head includes Normalization layer, Pooling layer and Fully Connected layer. $\textbf{Y1}$ is input to the dynamic face detection Task-Specific Head to determine the position of the face in each video frame, represented by the bounding box $\textbf{B} \in {R}^{16 \times 4}$. $\textbf{Y2}$ is input to the dynamic face landmark Task-Specific Head to obtain the five points $\textbf{P} \in {R}^{16 \times 10}$ (left eye, right eye, nose, left mouth corner, and right mouth corner) in each video frame. $\textbf{Y3}$ is input to the DFER Task-Specific Head to calculate the probability of each facial expression $\textbf{E} \in {R}^{1 \times 7}$ (assuming there are 7 facial expression classes) in each video. The mathematical expressions for the above process are as follows:
\begin{equation}
\textbf{B} = \text{FC}(\text{Pool}(\text{Norm}(\textbf{Y1})))
\label{eq:fc1}
\end{equation}
\begin{equation}
\textbf{P} = \text{FC}(\text{Pool}(\text{Norm}(\textbf{Y2})))
\label{eq:fc2}
\end{equation}
\begin{equation}
\textbf{E} = \text{FC}(\text{Pool}(\text{Norm}(\textbf{Y3})))
\label{eq:fc3}
\end{equation}

\section{EXPERIMENTS}

\subsection{Datasets}

RAVDESS \cite{c23}: The Ryerson Audio-Visual Database of Emotional Speech and Song is a dataset composed of emotional performances by 24 North American professional actors, recorded in a laboratory environment with North American accent. The facial emotions described in this dataset include anger, disgust, fear, happiness, neutral, sadness, and surprise. Each emotion includes two intensity variations: normal and strong. In this work, we only use the visual dataset, which contains 1,440 video files. Among them, there are 288 videos for neutral emotions, and 192 videos for each of the other emotions. Our model evaluation is conducted using 5-fold cross-validation on subjects-independent of the emotion.

CREMA-D \cite{c24}: The Crowd-sourced Emotional Multimodal Actor Dataset contains emotional performances by 91 professional actors from around the world, representing various countries and ethnicities. These actors performed emotional expressions in the laboratory environment with different accents. Each actor displays six different emotions: anger, disgust, fear, happiness, neutral and sadness. In addition, each emotion was expressed at low, medium, high, and unspecified intensity levels during dialogues. The dataset has a total of 7,442 video clips, which 1,087 video clips depicting neutral emotions, and 1,271 video clips for each of the other emotions. The model evaluation adopts 5-fold cross-validation with subjects-independent of the emotion.

MEAD \cite{c25}: The Multi-view Emotional Audio-visual Dataset is a video corpus of talking faces from 60 actors, who speak with eight different emotions (anger, disgust, contempt, fear, happiness, sadness, surprise, and neutral) at three different intensity levels (except neutral). The videos were recorded simultaneously from seven different angles under strictly controlled environment to capture high-quality facial expression details. According to the publicly available part of this dataset, we used data from 48 actors, comprising a total of 6,568 video clips. Among these, 380 clips depict neutral emotions, while each of the other emotions has 884 clips. The model evaluation still uses 5-fold cross-validation with subjects-independent of the emotion.

\begin{figure}[t]
  \centering
  \includegraphics[scale=0.56]{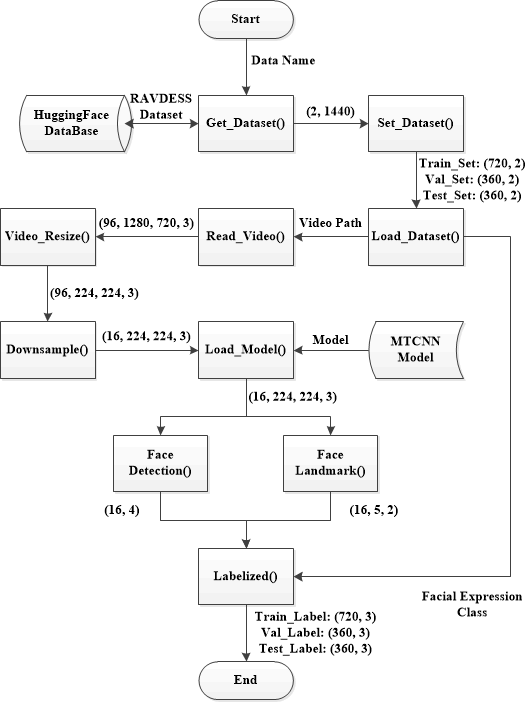}
  \caption{Multi-Task Label Data Preprocessing Flow-Chart}
  \label{fig:MTflowchart}
\end{figure}

\subsection{Implementation Details}

Preprocessing: In this study, our main preprocessing work is to perform multi-task labeling on various dataset. Since the various dataset used don't include face position and face landmark labels, we employ the MTCNN \cite{c26} model to perform frame-level annotations on the video data, leveraging its face detection and landmark functions to collect multi-task data labels. Using the RAVDESS \cite{c23} dataset as an example, it consists of two sub-datasets: video path data and emotion label data. The video path data is input into the Read\_Video() function, allowing us to retrieve video information, especially the total number of frames, resolution, and the number of channels. To meet the input requirements of the VideoMAE \cite{c19} encoder model, we preprocess these video datasets by adjusting the resolution of each frame to $224 \times 224 $ using the image scaling function. 

Additionally, since the maximum input is 16 frames of video, we process the total frame count of the source videos using average downsampling to ensure that all videos have exactly 16 frames. Next, we load the MTCNN \cite{c26} model to perform face detection and face landmark on the preprocessed videos $ \textbf{X} \in {R}^{16 \times 224 \times 224 \times 3} $, obtaining frame-level face bounding box annotations $16 \times 4 $ and 5-point annotations $16 \times 5 \times 2 $. Finally, we summarize and categorize the emotion label data to complete the multi-task labeling data preprocessing. The data preprocessing process is illustrated in Fig. 3.

Fine-tuning: After preprocessing the video dataset to obtain data that meets the input requirements of the VideoMAE \cite{c19} encoder, we use the pre-trained model based on the ViT-L \cite{c22} from MAE-DFER \cite{c16}. As described in Section 3.2, to enhance global-local feature interactions across tasks, the 5-layer, 3-head ViT Decoder cascaded module is configured for each task to associate global and local representation information. Additionally, MTL is employed to handle each task, particularly using task-specific heads equipped with Pooling layer and Fully Connected layer for fine-tuning on downstream tasks. The AdamW \cite{c27} optimizer is used with the base learning rate of 1e-3 and the weight decay of 0.001. Mean squared error (MSE) is used as the loss function for dynamic face detection and landmark, with the loss weight of 0.5 for each task. Sparse Categorical Cross-Entropy is used as the loss function for DFER, with the loss weight of 1.5, and Accuracy is used as the training evaluation metric. The DFER evaluation metrics for the model include Unweighted Average Recall (UAR) and Weighted Average Recall (WAR).


\subsection{Structure strategy}

To demonstrate the effectiveness of the multi-task cascaded network in global-local feature interaction, we propose six structure strategies of the network framework to analyze various architectures from STL to MTL. The following structure strategies are extensions based on Fig. 1.

\begin{itemize}[itemsep=0em,topsep=0em]
  \item \textbf{First Architecture:} The STL network, consisting of the VideoMAE \cite{c19} encoder and the task-specific head with only Fully Connected layer.
  \item \textbf{Second and Third Architectures:} Non-Fully Shared MTL networks, consisting of the VideoMAE \cite{c19} encoder and multiple task-specific decoders. In the second architecture, all task-specific decoders are MLP, while in the third architecture, all task-specific decoders are our proposed ViT Decoder.
  \item \textbf{Fourth and Fifth Architectures:} Fully Shared MTL networks, consisting of the VideoMAE \cite{c19} encoder, a single shared decoder, and multiple task-specific decoders. In the fourth architecture, both the shared decoder and task-specific decoders are MLP. In the fifth architecture, both the shared decoder and task-specific decoders are our proposed ViT Decoder.
  \item \textbf{Sixth Architecture:} Our proposed multi-task cascaded network, which consists of the VideoMAE \cite{c19} encoder and multiple cascaded task-specific decoders, where the task-specific decoders are ViT Decoder.
\end{itemize}

\subsection{Results}

\subsubsection{Ablation Study}

\begin{table}
\caption{Ablation study on the structure strategies. MTL: Multi-task learning method or not. Decoder: MLP or ViT Decoder. UAR: unweighted average recall. WAR: weighted average recall. BOLD: The best results.}
\label{table_ablation}
\begin{center}
\resizebox{\linewidth}{!}{
\begin{tabular}{lccccc}
\toprule
Dataset & MTL & Type & Decoder & UAR & WAR \\
\midrule
\multirow{6}{*}{RAVDESS \cite{c23}}
& $\times$     & $\textbf{-}$         & $\textbf{-}$      &  $72.36$             & $74.44$ \\
& \checkmark   & Non-Fully Shared     & MLP               &  $75.56$             & $77.28$ \\
& \checkmark   & Non-Fully Shared     & ViT Decoder       &  $81.11$             & $82.17$ \\
& \checkmark   & Fully Shared         & MLP               &  $75.95$             & $76.84$ \\
& \checkmark   & Fully Shared         & ViT Decoder       &  $80.02$             & $81.43$ \\
& \checkmark   & Cascaded             & ViT Decoder       &  $\textbf{82.73}$    & $\textbf{83.69}$
\\
\midrule
\multirow{6}{*}{CREMA-D \cite{c24}}
& $\times$     & $\textbf{-}$         & $\textbf{-}$      &  $75.87$             & $76.21$ \\
& \checkmark   & Non-Fully Shared     & MLP               &  $77.23$             & $79.89$ \\
& \checkmark   & Non-Fully Shared     & ViT Decoder       &  $83.40$             & $83.67$ \\
& \checkmark   & Fully Shared         & MLP               &  $76.25$             & $77.55$ \\
& \checkmark   & Fully Shared         & ViT Decoder       &  $81.64$             & $82.38$ \\
& \checkmark   & Cascaded             & ViT Decoder       &  $\textbf{84.71}$    & $\textbf{85.03}$
\\
\midrule
\multirow{6}{*}{MEAD \cite{c25}}
& $\times$     & $\textbf{-}$         & $\textbf{-}$      &  $77.83$             & $79.61$ \\
& \checkmark   & Non-Fully Shared     & MLP               &  $81.40$             & $82.14$ \\
& \checkmark   & Non-Fully Shared     & ViT Decoder       &  $84.75$             & $86.57$ \\
& \checkmark   & Fully Shared         & MLP               &  $79.12$             & $80.93$ \\
& \checkmark   & Fully Shared         & ViT Decoder       &  $84.03$             & $84.74$ \\
& \checkmark   & Cascaded             & ViT Decoder       &  $\textbf{87.51}$    & $\textbf{88.44}$
\\
\bottomrule
\end{tabular}
}
\end{center}
\end{table}

\begin{table}[b!]
\caption{Comparison with state-of-the-art methods on RAVD-ESS (7-class). SSL: self-supervised learning method or not. MTL: Multi-task learning method or not. UAR: unweighted average recall. WAR: weighted average recall.}
\label{table_ravdess}
\begin{center}
\resizebox{\linewidth}{!}{
\begin{tabular}{lcccc}
\toprule
Method & SSL & MTL & UAR & WAR \\
\midrule
VO-LSTM \cite{c28}         & $\times$       & $\times$      & $\textbf{-}$          & $60.50$ \\
3D ResNeXt-50 \cite{c29}   & $\times$       & $\times$      & $\textbf{-}$          & $62.99$ \\
SVFAP \cite{c30}           & \checkmark     & $\times$      & $75.15$               & $75.01$ \\
MAE-DFER \cite{c16}        & \checkmark     & $\times$      & $\underline{75.91}$   & $\underline{75.56}$ \\
\midrule
MNC \cite{c18}             & $\times$       & \checkmark    & $61.81$               & $61.43$ \\
MTL-ER \cite{C31}          & $\times$       & \checkmark    & $73.54$               & $74.19$ \\
MTFormer \cite{c17}        & \checkmark     & \checkmark    & $\underline{79.40}$   & $\underline{79.68}$ \\
\midrule
MTCAE-DFER (ours)          & \checkmark     & \checkmark    & $\textbf{82.73}$      & $\textbf{83.69}$ \\
\bottomrule
\end{tabular}
}
\end{center}
\end{table}

According to the structure strategy in Section 4.3, ablation experiments are conducted on STL and various MTL with different decoder types. We first start with the analysis of the RAVDESS \cite{c23} dataset. As shown in Table 1, the ablation results of the six architectures show that compared to STL, the various MTL strategies improve WAR by at least 2.4\% (76.84\% vs. 74.44\%). Notably, the cascaded structure further improves WAR by 9.25\% (83.69\% vs. 74.44\%). Furthermore, among the MTL strategies, architectures with the ViT Decoder module outperform those with MLP decoder. Specifically, the performance difference in WAR for the Non-Fully Shared type is 4.89\% (82.17\% vs. 77.28\%), while the performance difference for the Fully Shared type is 4.59\% (81.43\% vs. 76.84\%) WAR. Finally, when employing the ViT Decoder module and further using the cascaded structure to handle tasks, WAR improves by 1.52\% (83.69\% vs. 82.17\%) compared with the Non-Fully Shared with the ViT Decoder, and 2.26\% (83.69\% vs. 81.43\%) WAR higher than the Fully Shared with the ViT Decoder.

Similarly, from the ablation experiment results for the CREMA-D \cite{c24} shown in Table 1, it can be observed that the performance of MTL structure better than STL structure, with the increased by at least 1.34\% (77.55\% vs. 76.21\%) in terms of WAR. The multi-task cascaded network further increases by 8.82\% (85.03\% vs. 76.21\%) WAR. Additionally, compared to the MTL with MLP decoder, the Non-Fully Shared MTL with ViT Decoder improves WAR performance by 3.78\% (83.67\% vs. 79.89\%), while the Fully Shared MTL with ViT Decoder improves performance by 4.83\% (82.38\% vs. 77.55\%) WAR. More importantly, the cascaded MTL with ViT Decoder not only outperforms the Non-Fully Shared MTL with ViT Decoder by 1.36\% (85.03\% vs. 83.67\%) WAR, but also exceeds the Fully Shared MTL with ViT Decoder by 2.65\% (85.03\% vs. 82.38\%) WAR.

Finally, as shown in Table 1 MEAD \cite{c25} ablation results, the MTL structure outperforms the STL structure, with the improved by at least 1.32\% (80.93\% vs. 79.61\%) in terms of WAR, while the multi-task cascaded network exceeds its by 8.83\% (88.44\% vs. 79.61\%) WAR. In the Non-Fully Shared MTL structure, the network with ViT Decoder achieves 4.43\% (86.57\% vs. 82.14\%) WAR better than the one with MLP. In the Fully Shared MTL structure, the network with ViT Decoder outperforms the one with MLP by 3.81\% (84.74\% vs. 80.93\%) WAR. Moreover, the multi-task cascaded network significantly outperforms both the Non-Fully Shared MTL with ViT Decoder and the Fully Shared MTL with ViT Decoder, increasing WAR by 1.87\% (88.44\% vs. 86.57\%) and 3.70\% (88.44\% vs. 84.74\%) respectively.

\subsubsection{Comparison with State-of-the-art Methods}

We first compared MTCAE-DFER with other previous SOTA models on the RAVDESS \cite{c23}. As shown in Table 2, the SOTA STL supervised model is 3D ResNeXt-50 \cite{c29}, which has WAR performance difference of 20.70\% (83.69\% vs. 62.99\%) with our MTCAE-DFER model. Compared to the SOTA STL self-supervised model, MTCAE-DFER outperforms the MAE-DFER \cite{c16} model by 8.13\% (83.69\% vs. 75.56\%) in terms of WAR. Furthermore, our proposed MTCAE-DFER model achieves performance by 9.50\% (83.69\% vs. 74.19\%) WAR over the SOTA MTL supervised model MTL-ER \cite{C31}. When compared to the SOTA MTL self-supervised model MTFormer \cite{c17}, the MTCAE-DFER model increases WAR by 4.01\% (83.69\% vs. 79.68\%).

The previous SOTA model results on CREMA-D \cite{c24} are presented in Table 3. We observed that the MTCAE-DFER model surpasses the 3D ResNeXt-50 \cite{c29} STL supervised model by 15.89\% (85.03\% vs. 69.14\%) in WAR. Compared to the STL self-supervised model MAE-DFER \cite{c16}, it improves performance by 7.65\% (85.03\% vs. 77.38\%) in terms of WAR. Additionally, in the field of MTL, the MTCAE-DFER model outperforms the MTL-ER \cite{C31} supervised learning model by 9.41\% (85.03\% vs. 75.62\%) WAR, while compared to the self-supervised learning model MTFormer \cite{c17}, it achieves WAR performance increase of 4.29\% (85.03\% vs. 80.74\%).

In Table 4, we compare the results of the current SOTA models on the MEAD \cite{c25}. The results show that the MTCAE-DFER outperforms both the STL supervised model 3D ResNeXt-50 \cite{c29} and the STL self-supervised model MAE-DFER \cite{c16}, increasing the WAR results by 15.88\% (88.44\% vs. 72.56\%) and 7.49\% (88.44\% vs. 80.95\%), respectively. Moreover, compared with the MTL model results, the MTCAE-DFER model improves WAR by 10.08\% (88.44\% vs. 78.36\%) over the MTL-ER \cite{C31} supervised model, and outperforms the MTFormer \cite{c17} self-supervised model by 3.7\% (88.44\% vs. 84.74\%) WAR.

\begin{table}
\caption{Comparison with state-of-the-art methods on CREMA-D (6-class). SSL: self-supervised learning method or not. MTL: Multi-task learning method or not. UAR: unweighted average recall. WAR: weighted average recall.}
\label{table_creamd}
\begin{center}
\resizebox{\linewidth}{!}{
\begin{tabular}{lcccc}
\toprule
Method & SSL & MTL & UAR & WAR \\
\midrule
VO-LSTM \cite{c28}         & $\times$       & $\times$      & $\textbf{-}$          & $66.80$ \\
3D ResNeXt-50 \cite{c29}   & $\times$       & $\times$      & $\textbf{-}$          & $69.14$ \\
SVFAP \cite{c30}           & \checkmark     & $\times$      & $77.31$               & $77.37$ \\
MAE-DFER \cite{c16}        & \checkmark     & $\times$      & $\underline{77.33}$   & $\underline{77.38}$ \\
\midrule
MNC \cite{c18}             & $\times$       & \checkmark    & $67.93$               & $68.57$ \\
MTL-ER \cite{C31}          & $\times$       & \checkmark    & $74.45$               & $75.62$ \\
MTFormer \cite{c17}        & \checkmark     & \checkmark    & $\underline{79.34}$   & $\underline{80.74}$ \\
\midrule
MTCAE-DFER (ours)          & \checkmark     & \checkmark    & $\textbf{84.71}$      & $\textbf{85.03}$ \\
\bottomrule
\end{tabular}
}
\end{center}
\end{table}

\begin{table}
\caption{Comparison with state-of-the-art methods on MEAD (8-class). SSL: self-supervised learning method or not. MTL: Multi-task learning method or not. UAR: unweighted average recall. WAR: weighted average recall.}
\label{table_mead}
\begin{center}
\resizebox{\linewidth}{!}{
\begin{tabular}{lcccc}
\toprule
Method & SSL & MTL & UAR & WAR \\
\midrule
VO-LSTM \cite{c28}         & $\times$       & $\times$      & $\textbf{-}$          & $68.87$ \\
3D ResNeXt-50 \cite{c29}   & $\times$       & $\times$      & $\textbf{-}$          & $72.56$ \\
SVFAP \cite{c30}           & \checkmark     & $\times$      & $80.04$               & $80.23$ \\
MAE-DFER \cite{c16}        & \checkmark     & $\times$      & $\underline{80.71}$   & $\underline{80.95}$ \\
\midrule
MNC \cite{c18}             & $\times$       & \checkmark    & $70.82$               & $70.16$ \\
MTL-ER \cite{C31}          & $\times$       & \checkmark    & $77.26$               & $78.36$ \\
MTFormer \cite{c17}        & \checkmark     & \checkmark    & $\underline{84.03}$   & $\underline{84.74}$ \\
\midrule
MTCAE-DFER (ours)          & \checkmark     & \checkmark    & $\textbf{87.51}$      & $\textbf{88.44}$ \\
\bottomrule
\end{tabular}
}
\end{center}
\end{table}

\subsection{Analysis and Reasoning}

Based on the above results, whether in STL or MTL, the results of multi-task cascaded learning MTCAE-DFER consistently surpass the current SOTA models across all datasets. This demonstrates that the Cascaded ViT Decoder module plays a crucial role in handling the cascading relationship between related tasks, which highlights the necessity of global and local feature interaction.

Considering different structure strategies, we observed in the ablation study that even when all using the ViT Decoder module, the performance of the MTL framework is inferior to the multi-task cascaded learning framework. This suggests that the cascaded network is effective in transmitting representational information between related tasks, while the cascaded approach strengthens the effectiveness of global-local dynamic feature interaction to a certain extent. Moreover, all with the same shared VideoMAE encoder, there is a noticeable performance difference between STL and MTL, further indicating that the MTL framework enhances the model's generalization ability.

Finally, in this work, utilizing the ViT Decoder module to cascade the three face-related tasks effectively explains that the local key features generated between related tasks guide the global dynamic features. This approach helps the main task focus on key features, while eliminating some redundant features. However, this framework demands significant computing resources, particularly due to the self-attention mechanism in the ViT architecture, which increases computational costs. On the other hand, this approach has certain limitations in practical applications, as it requires related tasks of the target task to form multiple tasks for cascade interaction. In this paper, face-related multi-tasks is adopted to map and address the problem of DFER utilizing multiple models to process facial data.

\section{CONCLUSIONS AND FUTURE WORKS}

\subsection{Conclusions}

This work is the further exploration of autoencoder-based MTL structure on cascaded network. We constructed the ViT Decoder module to address the inter-task relationships, enhancing global-local feature interactions across related tasks. Based on this module, we propose a new multi-task cascaded learning framework to improve the accuracy of DFER task. This paper studies the performance of MTL network structure under the same module configuration (shared VideoMAE \cite{c19} encoder, and multiple task-specific decoders, either ViT Decoder or MLP). By exploring six different structure strategies, we examined the robustness of the multi-task cascaded network and the effectiveness of global-local dynamic feature interaction. 

The results show that, whether ablation study or comparisons with SOTA model, the Cascaded ViT Decoder module enhances the generalization ability of the model in MTL networks. Moreover, this network architecture provides insight into the local dynamic feature generated from dynamic face detection and dynamic face landmark tasks, which helps identify key dynamic features necessary for DFER within the global representation information space.

\subsection{Future Works}

In future work, this framework has a certain versatility and can be applied to various related tasks to strengthen global-local feature interaction across tasks. Specifically, it could be combined with current emotion-based text generation tasks (video content description, video subject emotion inference, and video subject emotion description) to understand the emotional changes of the subject in the video, while better evaluate the model's ability to extract dynamic features within video understanding. 

On the other hand, this work can be extended to multi-modal model by integrating a unified multi-modal feature encoder to process multi-modal information, thereby obtaining global multi-modal features. The multi-task cascaded network uses these features to further improve the accuracy of each task.


{\small
\bibliographystyle{ieee}
\bibliography{egbib}
}

\end{document}